# MASF-YOLO: An Improved YOLOv11 Network for Small Object Detection on Drone View

Liugang Lu, Dabin He, Congxiang Liu, Zhixiang Deng

*Abstract*— **With the rapid advancement of Unmanned Aerial Vehicle (UAV) and computer vision technologies, object detection from UAV perspectives has emerged as a prominent research area. However, challenges for detection brought by the extremely small proportion of target pixels, significant scale variations of objects, and complex background information in UAV images have greatly limited the practical applications of UAV. To address these challenges, we propose a novel object detection network Multi-scale Context Aggregation and Scale-adaptive Fusion YOLO (MASF-YOLO), which is developed based on YOLOv11. Firstly, to tackle the difficulty of detecting small objects in UAV images, we design a Multi-scale Feature Aggregation Module (MFAM), which significantly improves the detection accuracy of small objects through parallel multi-scale convolutions and feature fusion. Secondly, to mitigate the interference of background noise, we propose an Improved Efficient Multi-scale Attention Module (IEMA), which enhances the focus on target regions through feature grouping, parallel sub-networks, and cross-spatial learning. Thirdly, we introduce a Dimension-Aware Selective Integration Module (DASI), which further enhances multi-scale feature fusion capabilities by adaptively weighting and fusing low-dimensional features and high-dimensional features. Finally, we conducted extensive performance evaluations of our proposed method on the VisDrone2019 dataset. Compared to YOLOv11-s, MASF-YOLO-s achieves improvements of 4.6% in mAP@0.5 and 3.5% in mAP@0.5:0.95 on the VisDrone2019 validation set. Remarkably, MASF-YOLO-s outperforms YOLOv11-m while requiring only approximately 60% of its parameters and 65% of its computational cost. Furthermore, comparative experiments with state-of-the-art detectors confirm that MASF-YOLO-s maintains a clear competitive advantage in both detection accuracy and model efficiency.**

## I. INTRODUCTION

In recent years, the rapid development of unmanned aerial vehicle (UAV) and deep learning technologies has brought revolutionary changes to multiple fields such as agriculture, emergency rescue, and traffic monitoring [1], [2], [3]. UAV have become essential tools for data collection and real-time decision-making, with their flexibility, cost-effectiveness, and ability to access hard-to-reach areas. Meanwhile, deep learning algorithms have shown superoiror over tranditional methods in processing complex remote sensing image data, enabling UAV to perform tasks such as object detection and segmentation with higher efficiency. The integration of UAV and deep learning offers innovative solutions to information collection across various domains.

Object detection through aerial vision constitutes a critical link in UAV missions. Currently, deep learning-based approaches for image object detection can be primarily categorized into two-stage object detection algorithms and single-stage object detection algorithms. Two-stage object detection algorithms achieve object detection through two phases: "region proposal" and "classification regression." Representative algorithms include Fast R-CNN [4], Faster R-CNN [5], and Mask R-CNN [6]. Their strengths lie in high accuracy and multi-task capabilities: through refined region proposals and classification regression. Their shortcoming include high computational complexity and poor real-time performance, making them less suitable for applications with high-speed requirements. In contrast, single-stage detection algorithms complete object classification and localization in a single forward pass, significantly improving detection speed. Representative single-stage detection algorithms include the YOLO series [7], [8], [9] and SSD [10]. YOLO achieves end-to-end detection by dividing the image into grids , where each grid is responsible for predicting the location and category of objects. In contrast, SSD employs multi-scale feature maps for detection, integrating an anchor mechanism to effectively address the detection requirements of objects at varying scales. Benefiting from their fast detection speed that enables real-time inference, single-stage detectors have been widely adopted as the preferred solution for UAV applications, autonomous driving, and other time-critical systems. However, further improvements are still required to enhance their performance in complex scenarios.

Despite the achievements of UAV remote sensing object detection in many tasks, it still faces numerous technical challenges. Firstly, due to the high shooting distance of UAV, most targets occupy an extremely small proportion of pixels in the images, making feature extraction difficult and easily leading to missed or false detections. Secondly, complex and varied noise in the images increases the difficulty of detection. Furthermore, due to the varying shooting angles of UAV, the significant scale-shape variation of objects in images further complicates detection. Besides, the light-weight and high speed of detection models are also required in UVA missions. Therefore, addressing these challenges is the key way for improving the efficiency and robustness of object detection by UAV. The investigations hold significant theoretical and practical value for UAV related application fields.

To address these challenges, this paper proposes a high-precision algorithm maintaining light-weight framework specifically designed for small object detection in UAV images, named MASF-YOLO. Through rigorous empirical analysis and experimental validation, we have systematically

Liugang Lu is with the College of Science, Sichuan Agricultural University, Ya'an 625000 , China (e-mail: lglu2025@163.com).

Dabin He is with the College of Science, Sichuan Agricultural University, Ya'an 625000 , China (e-mail: dabinhe@163.com).

Congxiang Liu is with the College of Science, Sichuan Agricultural University, Ya'an 625000 , China (e-mail: liucongxiang@163.com).

Zhixiang Deng is with the College of Science, Sichuan Agricultural University, Ya'an 625000 , China (e-mail: 15181178952 @163.com).

demonstrated the effectiveness of multiple innovative design concepts incorporated in our method for this specific task.

More detailed, the novelty and contributions of our work can be listed as follows:

- To address the issue of small objects losing detailed information due to repeated downsampling, we construct a high-resolution small object detection layer. This architecture incorporates P2-level fine-grained feature maps to fully leverage their preserved rich spatial details, significantly enhancing the model's feature representation capability for small-scale targets. Additionally, skip connections are added to the neck network to retain more shallow semantic information, which effectively mitigates semantic information loss in deep networks.
- In small object detection tasks, targets usually contain limited pixel information, so richer contextual information is needed to assist detection. To address this challenge, we propose a new multi-scale feature aggregation module (MFAM) that effectively captures rich contextual information of the target. This architecture achieves more effective feature extraction, leading significantly improvement of the detection accuracy of small objects.
- Background noise has consistently been a critical factor impairing object detection performance in UAV applications. To effectively suppress such interference, we propose an Improved Efficient Multi-scale Attention Module (IEMA), where feature interaction and enhancement are achieved through feature grouping, parallel subnetworks, and cross-spatial learning. It effectively improves target region feature representation meanwhile significantly suppresses background noise interference, thereby improves object detection performance in complex scenarios.
- To overcome the multi-scale feature fusion challenge in UAV-based small object detection, we introduce Dimension-Aware Selective Integration Module (DASI) to adaptively fuse low-dimensional features and high-dimensional features. It substantially improves the neck network's multi-scale representation capability for enhanced detection performance.

## II. Related work

### A. UAV Remote Sensing Object Detection

Unlike traditional images, remote sensing images are typically captured from a top-down perspective, resulting in targets characterized by arbitrary orientations and significant scale variations. These characteristics render traditional object detection methods designed for traditional images less effective in processing remote sensing images. To address these limitations, researchers have improved these methods from various perspectives to better suit the unique properties of remote sensing images. To address scale variations, LSKNet [11] introduces a large selective kernel mechanism to dynamically adjust the spatial receptive field, enabling better modeling of target contextual information. Meanwhile, PKINet [12] employs multi-scale convolutional kernels to extract local features of targets at different scales and incorporates a Contextual Anchor Attention (CAA) module to capture long-range contextual information, thereby enhancing the model's adaptability to scale variations. For small object detection, Chen et al. [13] propose a High-Resolution Feature Pyramid Network (HR-FPN) to improve the detection accuracy of small-scale targets while avoiding feature redundancy. To mitigate background interference, FFCA-YOLO [14] constructs a Spatial Context Awareness Module (SCAM) to model the global context of targets, thereby suppressing irrelevant background information and highlighting target features.

### B. Context Feature Representation

In computer vision tasks, objects in images are closely related to their surrounding environment. Appropriate context feature representation can effectively model both local and global information, thereby enhancing the detection capability of the model. To capture long-range dependencies while avoiding excessive computational overhead, Guo et al. [15] decomposed large-kernel convolutions and propose a linear attention mechanism, achieving a balance between network performance and computational cost. Ouyang et al. [16] designed an Efficient Multi-scale Attention (EMA) module, which effectively establishes short-term and long-term dependencies, enhancing the model's ability to capture multi-scale contextual information. Furthermore, considering the limitations of single-scale features in modeling contextual information, Xu et al. [17] propose a Multi-Dilated Channel Refiner (MDCR) module, which captures spatial features of different receptive field sizes by designing multi-dilation-rate convolutional layers, improving the model's multi-granularity semantic representation capability.

### C. Multi-Scale Feature Fusion

As one of the significant milestones in the field of object detection, Feature Pyramid Network (FPN) [18] pioneered multi-scale feature fusion through top-down pathways. Building upon FPN, PAFPN [19] introduces an additional bottom-up pathway, enabling better transmission of detailed information from lower layers. Furthermore, BiFPN [20] incorporates learnable weights to perform weighted fusion of different input features, allowing the network to learn the importance of each feature and achieve efficient feature integration. Additionally, Asymptotic Feature Pyramid Network(AFPN) [21] adopts a progressive approach to gradually fuse features from different levels, avoiding semantic gaps between non-adjacent levels.

## III. Proposed method

This section will elaborate on the proposed MASF-YOLO. The overall architecture of the MASF-YOLO network is illustrated in Figure 1. Specifically, we add a small object detection layer (P2 Layer) to the baseline, enabling the network to focus on detecting small objects. Secondly, considering the impact of target scale variations, we design a Multi-scale Feature Aggregation Module (MFAM) by optimizing PKINet [12]. This feature aggregation approach helps the backbone network capture rich contextual information, thereby improving the network's performance in detecting small objects. Furthermore, To enhance feature propagation and preserve fine-grained details, we incorporate cross-layer skip connections between shallow and deep feature maps in the neck network, termed Fusion. These Fusion establish direct pathways for transferring high-resolution spatial information from early layers, effectively compensating for the semantic information loss caused by deep network operations. Additionally, to mitigate the interference of background noise, we propose an Improved Efficient Multi-scale Attention (IEMA) module inspired by EMA [16]. This attention mechanism achieves feature interaction and enhancement through feature grouping, parallel sub-networks, and cross-spatial learning, effectively addressing the challenges posed by background noise. Finally, we introduce the Dimension-Aware Selective Integration (DASI) [17] module to enhance the multi-scale feature fusion capability of the neck network. This fusion mechanism adaptively aggregates low-dimensional and high-dimensional features, playing a crucial role in improving the detection accuracy of the network.

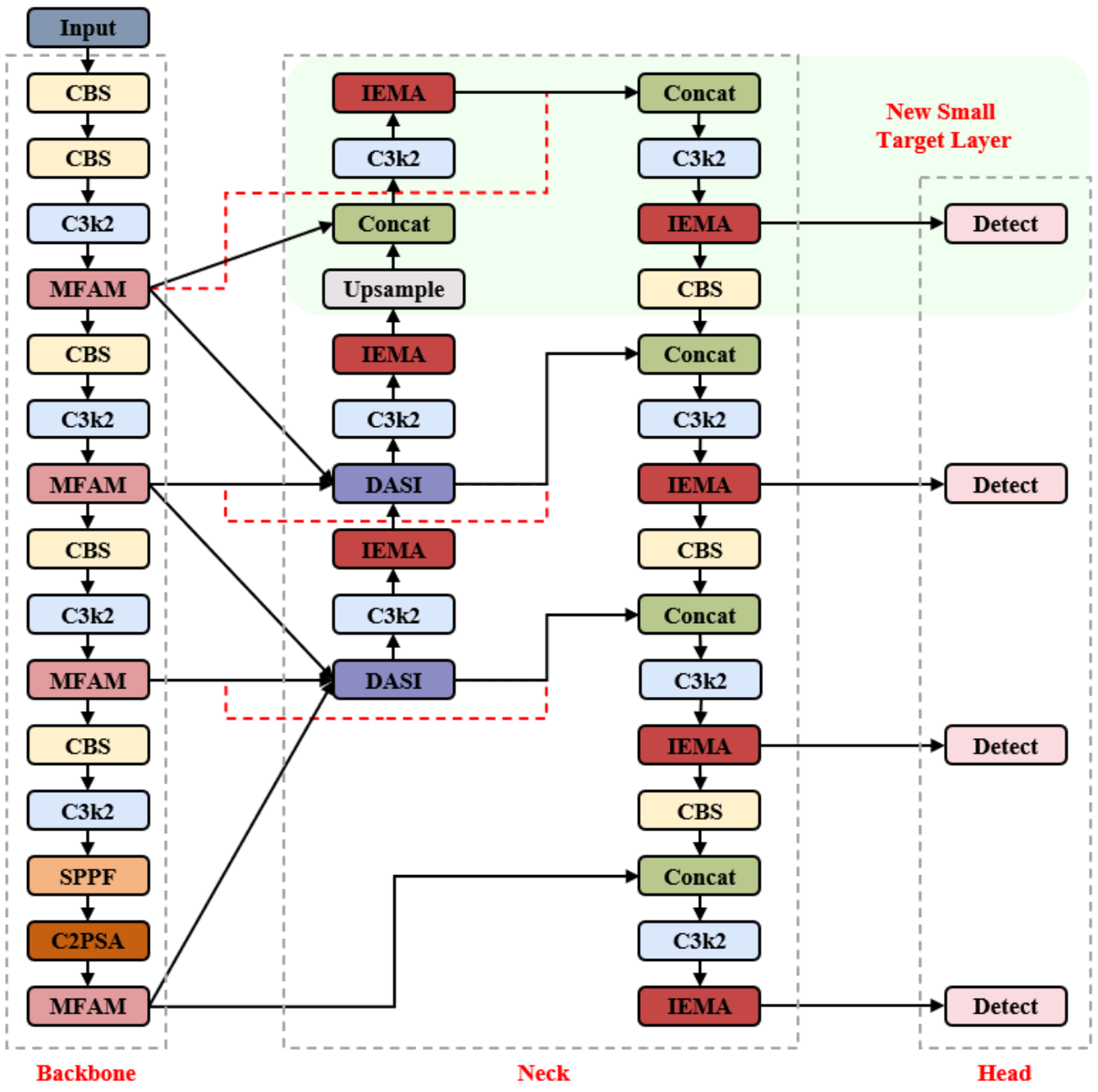

Figure 1. Overall framework of MASF-YOLO

## A. Multi-scale Feature Aggregation Module (MFAM)

Unlike general object detection, the significant scale variation of targets poses a substantial challenge in remote sensing object detection. Specifically, the backbone stage extracts limited effective semantic information, making it difficult to distinguish small objects from the background. To address this challenge, we propose the MFAM to capture rich contextual information of targets, enhancing the backbone's ability to extract features of small objects. The overall structure of MFAM is illustrated in Figure 2, which builds upon PKINet [12] with optimized design principles. The difference lies in the fact that the MFAM module utilizes two strip convolutions $1 \times k$ and $k \times 1$, to achieve an effect similar to that of a large-kernel convolution $k \times k$ ($k = 7, 9$), while removing the large-kernel convolution $11 \times 11$, significantly reducing computational costs. Additionally, the $3 \times 3$ convolution is modified to operate in parallel, further enhancing the representation of multi-scale features and avoiding the potential loss of small object semantic information that could arise from serial connections. The mathematical expression of MFAM can be written as:

$$Y_1 = DWConv_{3\times3}(X) \tag{1}$$

$$Y_2 = DWConv_{5\times5}(X) \tag{2}$$

$$Y_3 = DWConv_{7\times1}(DWConv_{1\times7}(X)) \tag{3}$$

$$Y_4 = DWConv_{9\times1}(DWConv_{1\times9}(X)) \tag{4}$$

$$Z = Y_1 \oplus Y_2 \oplus Y_3 \oplus Y_4 \oplus X \tag{5}$$

$$W = Conv_{1\times1}(Conv_{1\times1}(Z) \oplus X) \tag{6}$$

Where $DWConv_{k\times k}$ represent a depthwise separable convolution operation with a kernel size of $k \times k$. $Conv_{1\times1}$ represent a standard convolution operation with a kernel size of $1 \times 1$. The symbol $\oplus$ indicates the element-wise addition operation of feature maps. $X$ is the input feature map. $Y_1, Y_2, Y_3$ and $Y_4$ represent the output feature maps obtained after applying depthwise separable convolution operations with four different kernel sizes. $Z$ is the feature map resulting from the element-wise summation of the multi-scale features $Y_1, Y_2, Y_3, Y_4$ and the input feature $X$. $W$ is the output feature of MFAM.

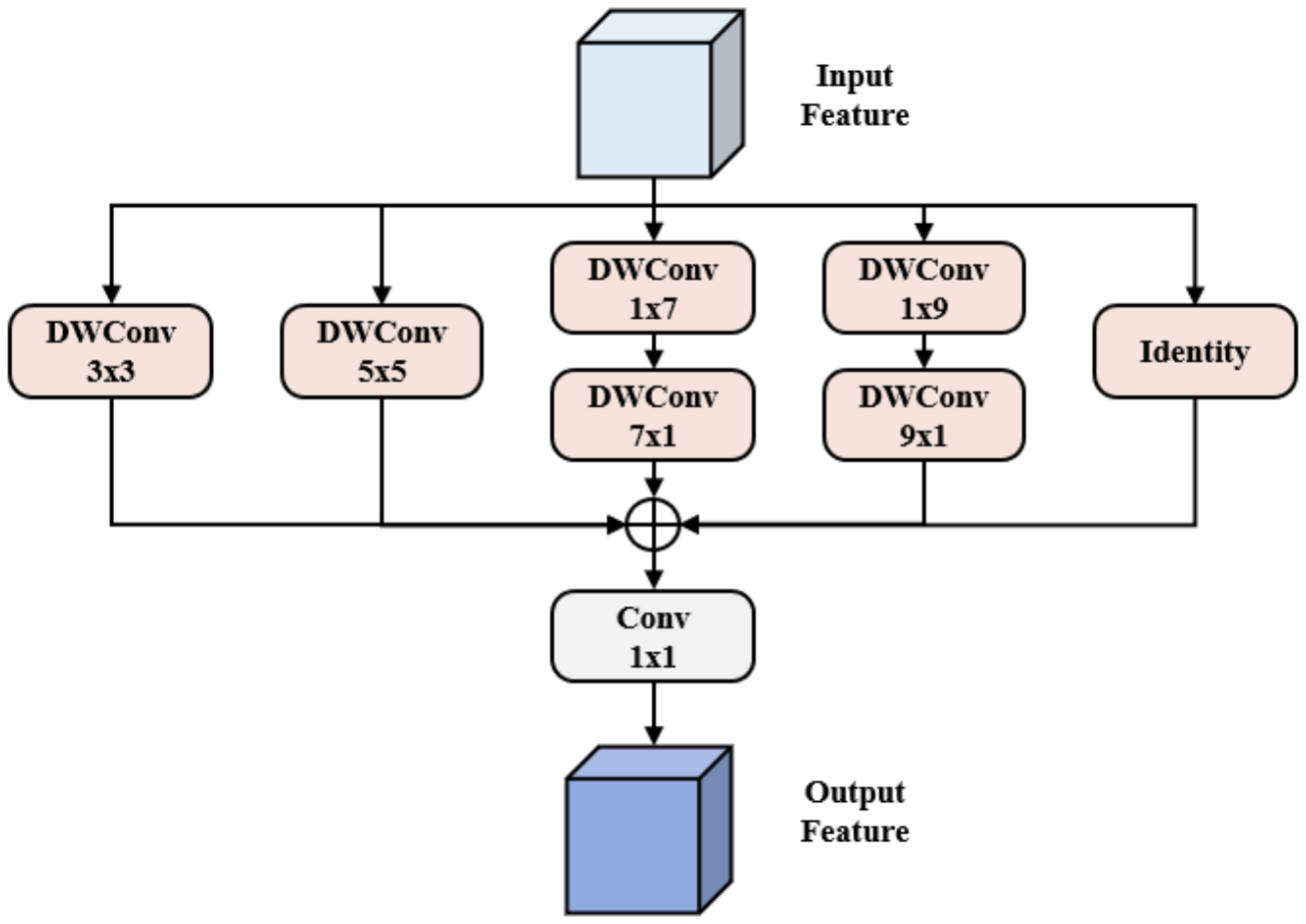

Figure 2. The structure of MFAM

Compared to the PKI Module [12], MKAM significantly enhances the detection capability of small targets by learning richer contextual features through multi-scale convolutions, while maintaining a more lightweight structure.

## B. Improved Efficient Multi-scale Attention(IEMA)

After passing through the MFAM in the backbone network, the feature maps already contain sufficient local contextual information. However, the influence of background noise still poses a significant challenge to the detection performance of the network. To address this challenge, it is necessary to effectively model the global relationship between targets and the background. Inspired by EMA [16] and InceptionNeXt [22], we construct IEMA module, as shown in Figure 3. Compared with EMA, IEMA primarily optimizes the local feature extraction component within the parallel subnetworks by introducing multi-scale depthwise separable convolutions, including $3 \times 3$, $1 \times 5$, and $5 \times 1$ kernels, along with an additional identity path. This optimization enhances directional feature extraction, enabling the model to more effectively capture multi-scale representations, thereby improving the modeling of global target-background relationships and strengthening the suppression of complex background interference. Meanwhile, IEMA retains EMA's global modeling capability through parallel subnetworks and cross-spatial learning mechanisms, facilitating feature interaction and enhancement.

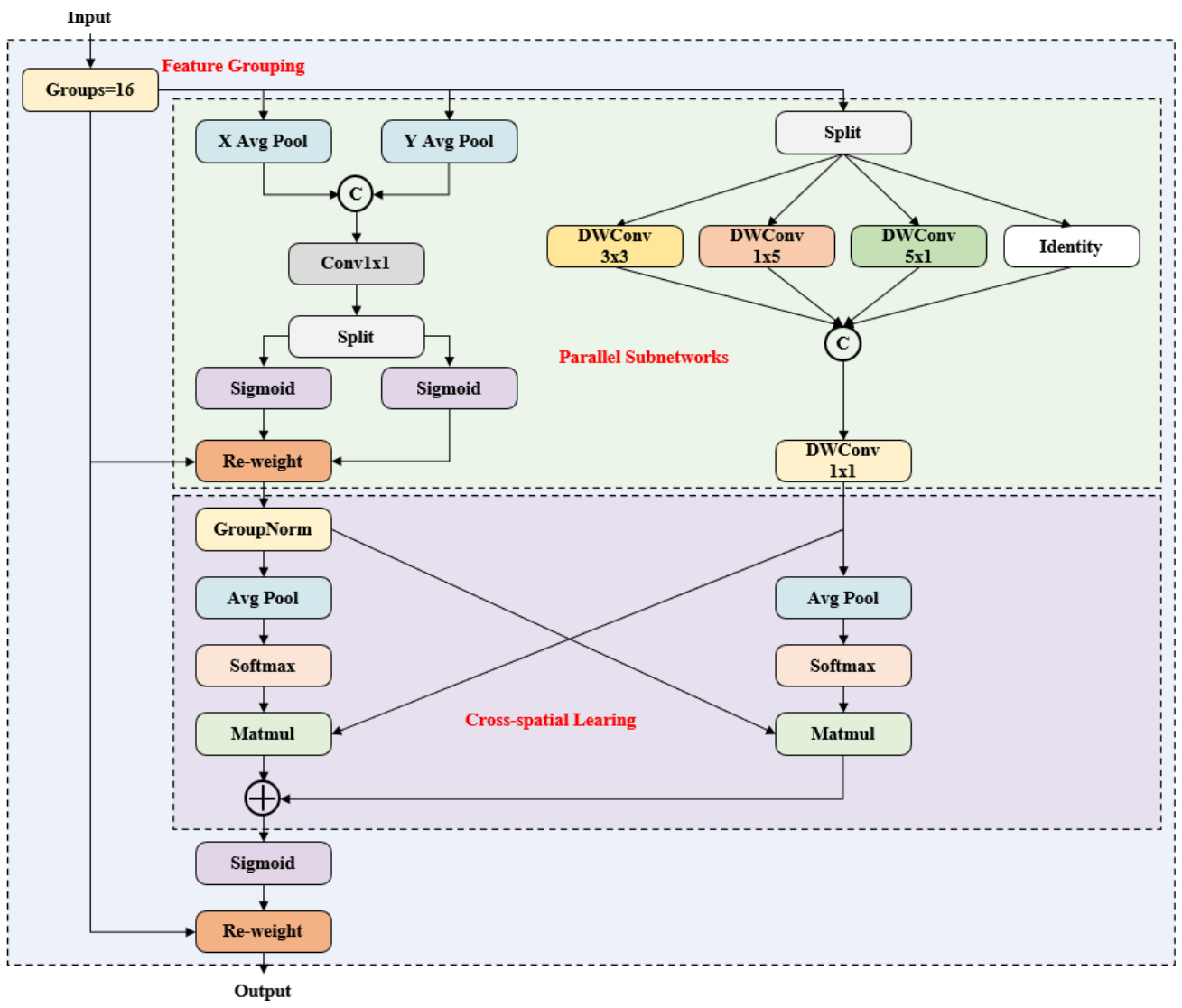

Figure 3. The structure of IEMA

### *C. Dimension-Aware Selective Integration Module(DASI)*

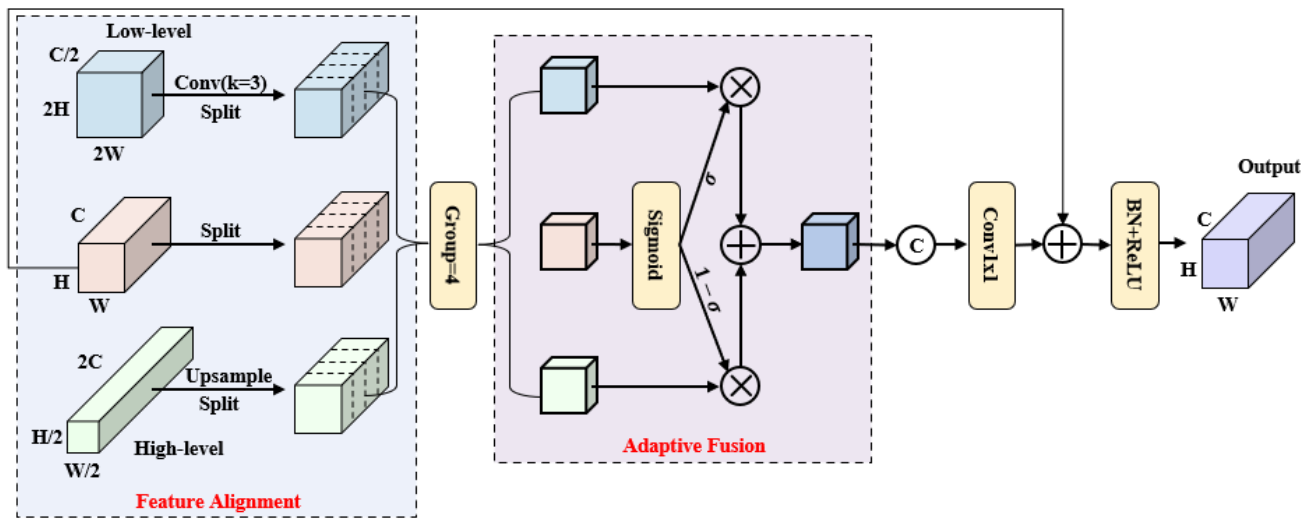

Figure 4. structure of DASI

In UAV remote sensing for small object detection, feature maps undergo multiple downsampling stages in the backbone network, causing high-level features to lose small object details while low-level features lack contextual information. Multi-scale feature aggregation effectively fuses semantic information across different levels, significantly improving detection accuracy for small objects. To address this issue, we introduce the Dimension-Aware Selective Integration Module (DASI) [17]. which adaptively fuses low-dimensional features and high-dimensional features. As shown in Figure 4, DASI first aligns low-dimensional and high-dimensional features with the current layer's features through operations such as convolution and interpolation. Then, it divides the three features into four equal parts along the channel dimension, ensuring that each part corresponds to a partition. Within each partition, the sigmoid activation function is used to obtain the activation values of the current layer's features, which are then used to weight and fuse the low-dimensional and high-dimensional features. Finally, the results from each partition are concatenated along the channel dimension, followed by a residual structure to preserve the semantic information of the current layer's features. By leveraging the current layer's features to adaptively fuse low-dimensional and high-dimensional features, DASI's multi-scale feature fusion mechanism not only improves the network's ability to detect small objects but also enhances its adaptability to complex remote sensing scenarios.

## IV. EXPERIMENTS

This section first introduces the datasets, experimental setup, training strategy, as well as the metrics used to evaluate the object detection performance of the model. Then using YOLOv11-s as the baseline and verifies the impact of each innovation on the baseline through ablation experiments. Furthermore, we compare our model with other state-of-the-art (SOTA) methods to demonstrate its competitive performance. To facilitate intuitive evaluation, we present comparative visualizations of detection results between our method and the baseline method, with these qualitative demonstrations showing strong consistency with quantitative metrics, collectively validating the effectiveness of our improvement strategy.

### *A. Dataset*

The VisDrone2019 dataset, developed by the AISKYEYE team from the Machine Learning and Data Mining Laboratory at Tianjin University, is a large-scale aerial image dataset comprising 288 video clips, 261,908 frames, and 10,209 static images captured by various drones in diverse scenarios. Covering 14 cities in China, the dataset encompasses both urban and rural environments, with annotations for multiple object categories such as pedestrians, cars, and bicycles. It includes images ranging from sparse to crowded scenes, as well as varying lighting and weather conditions. Due to its characteristics of abundant small objects, object overlaps, and complex backgrounds, the detection tasks are highly challenging. This dataset provides high-quality experimental resources for research on object detection and tracking from UAV perspectives, holding significant academic and practical value.

### *B. Training Sets*

The proposed model in this paper is implemented in PyTorch, with CUDA version 11.3, and the experimental environment includes the operating system Ubuntu 20.04 and an NVIDIA GeForce RTX 4090D 24G graphics card. The Stochastic Gradient Descent (SGD) optimizer is used for model training. The initial learning rate is set to 0.01, the momentum is 0.937, and the learning rate is dynamically adjusted using a cosine annealing strategy. The batch size during the training phase is set to 12, and the number of epochs is set to 100. Additionally, all images are resized to 640x640 pixels during the training phase.

### *C. Evaluation Metrics*

To comprehensively evaluate the performance of our proposed model, we employed several key metrics commonly used in object detection tasks: precision (P), recall (R), mAP@0.5, mAP@0.5:0.95, parameters (Params) and GFLOPs. This section outlines the formulas for calculating these metrics.

TABLE I. ABLATION EXPERIMENTAL RESULTS ON THE VISDRONE2019 VALIDATION DATASET

| Model | P2 Layer | MFAM | Fusion | IEMA | DASI | P | R | mAP50 (%) | mAP50:95 (%) | Params (M) |
|---|---|---|---|---|---|---|---|---|---|---|
| Baseline | | | | | | 0.572 | 0.312 | 0.446 | 0.294 | **9.42** |
| - | √ | | | | | 0.555 | 0.366 | 0.468 | 0.307 | 9.62 |
| - | √ | √ | | | | **0.578** | 0.379 | 0.478 | 0.319 | 10.93 |
| - | √ | √ | √ | | | 0.563 | 0.393 | 0.483 | 0.321 | 11.00 |
| - | √ | √ | √ | √ | | 0.562 | 0.402 | 0.488 | 0.324 | 11.01 |
| MASF-YOLO | √ | √ | √ | √ | √ | 0.563 | **0.407** | **0.492** | **0.329** | 12.05 |

TABLE II. PERFORMANCE COMPARISON BETWEEN ALL VERSIONS OF YOLOv11 AND MASF-YOLO ON THE VISDRONE2019 DATASET

| Baseline | $mAP_{50}^{val}$(%) | $mAP_{50:95}^{val}$(%) | $mAP_{50}^{test}$(%) | $mAP_{50:95}^{test}$(%) | Params(M) | GFLOPs |
|---|---|---|---|---|---|---|
| YOLOv11-n | 39.5 | 25.5 | 33.7 | 20.5 | 2.58 | 6.3 |
| MASF-YOLO-n | **43.2** | **28.2** | **37.3** | **23.0** | 3.31 | 14.4 |
| YOLOv11-s | 44.6 | 29.4 | 38.4 | 24.0 | 9.42 | 21.3 |
| MASF-YOLO-s | **49.2** | **32.9** | **42.8** | **26.8** | 12.05 | 44.3 |
| YOLOv11-m | 47.8 | 32.2 | 42.1 | 26.6 | 20.04 | 67.7 |
| MASF-YOLO-m | **52.3** | **35.1** | **45.4** | **28.6** | 27.56 | 143.9 |
| YOLOv11-l | 48.9 | 32.9 | 43.0 | 27.4 | 25.29 | 86.6 |

Precision is the ratio of correctly predicted positive instances (TP) to all instances predicted as positive (the sum of TP and FP). TP represents the number of true positives correctly identified, while FP represents the number of false positives incorrectly identified as positives. The formula for precision is as follows:

$$P = \frac{TP}{TP+FP} \tag{7}$$

Recall is the ratio of correctly predicted positive instances (TP) to all actual positive instances (the sum of TP and FN). FN represents the number of false negatives incorrectly identified as negatives. The formula for recall is as follows:

$$R = \frac{TP}{TP+FN} \tag{8}$$

mAP (mean Average Precision) is the average of AP (Average Precision) across all categories. With Intersection over Union (IoU) set to a constant value, the average precision for a category $(AP_i, i = 1,2,\dots,n)$ is the area under the Precision-Recall (P-R) curve. The formulas for AP and mAP are as follows:

$$AP = \int PRd(R) \tag{9}$$

$$mAP = \sum_{i=1}^{n} AP_i \tag{10}$$

Here, mAP@0.5 is obtained by calculating the mAP at an IoU threshold of 0.5, while mAP@0.5:0.95 is calculated by averaging the mAP values at IoU thresholds ranging from 0.5 to 0.95 with a step size of 0.05.

### D. *Ablation Study*

To validate the effectiveness of the proposed model in this paper, we selected YOLOv11-s as the baseline network and evaluated the impact of the P2 Layer, MFAM, Fusion, IEMA and DASI modules on the baseline network through ablation experiments. As shown in TABLE I, when each module was added to the baseline, most performance metrics exhibited an increasing trend. Therefore, these ablation experiments validate the effectiveness of the proposed method in this paper.

As shown in TABLE II, by adjusting the depth and width of the network, we evaluated different model sizes of MASF-YOLO and YOLOv11 on the VisDrone2019 validation set and test sets. It is evident that our proposed improvement strategies achieve optimal performance across all versions. Surprisingly, after applying our contributions to YOLOv11-s, its performance even surpasses that of YOLOv11-m, demonstrating superior accuracy-efficiency trade-offs in drone scenarios.

### E. *Comparison With State-of-the-Arts*

As shown in TABLE III, compared to state-of-the-art object detectors, the proposed model maintains excellent accuracy and demonstrates strong competitiveness. Additionally, in Figure 5, we present two highly representative detection results, where small targets missed by the baseline model (but successfully detected by MASF-YOLO-s are highlighted with red bounding boxes. It can be observed that MASF-YOLO-s achieves significantly more accurate detection.

TABLE III. COMPARISON RESULTS OF DIFFERENT OBJECT DETECTORS ON THEVISDRONE2019 VALIDATION DATASETS

| Model | mAP50 (%) | mAP50:95 (%) | Params (M) |
|---|---|---|---|
| Faster R-CNN[5] | 23.6 | 13.4 | 41.39 |
| Cascade R-CNN[23] | 24.7 | 14.5 | 69.18 |
| EfficientDet[20] | 26.0 | 14.1 | 18.53 |
| TPH-YOLOv5-s[24] | 37.4 | 21.7 | - |
| YOLOv8-m[8] | 43.4 | 26.5 | 25.90 |
| MSFE-YOLO-l[25] | 46.8 | 29.0 | - |
| MASF-YOLO-s | **49.2** | **32.9** | **12.05** |

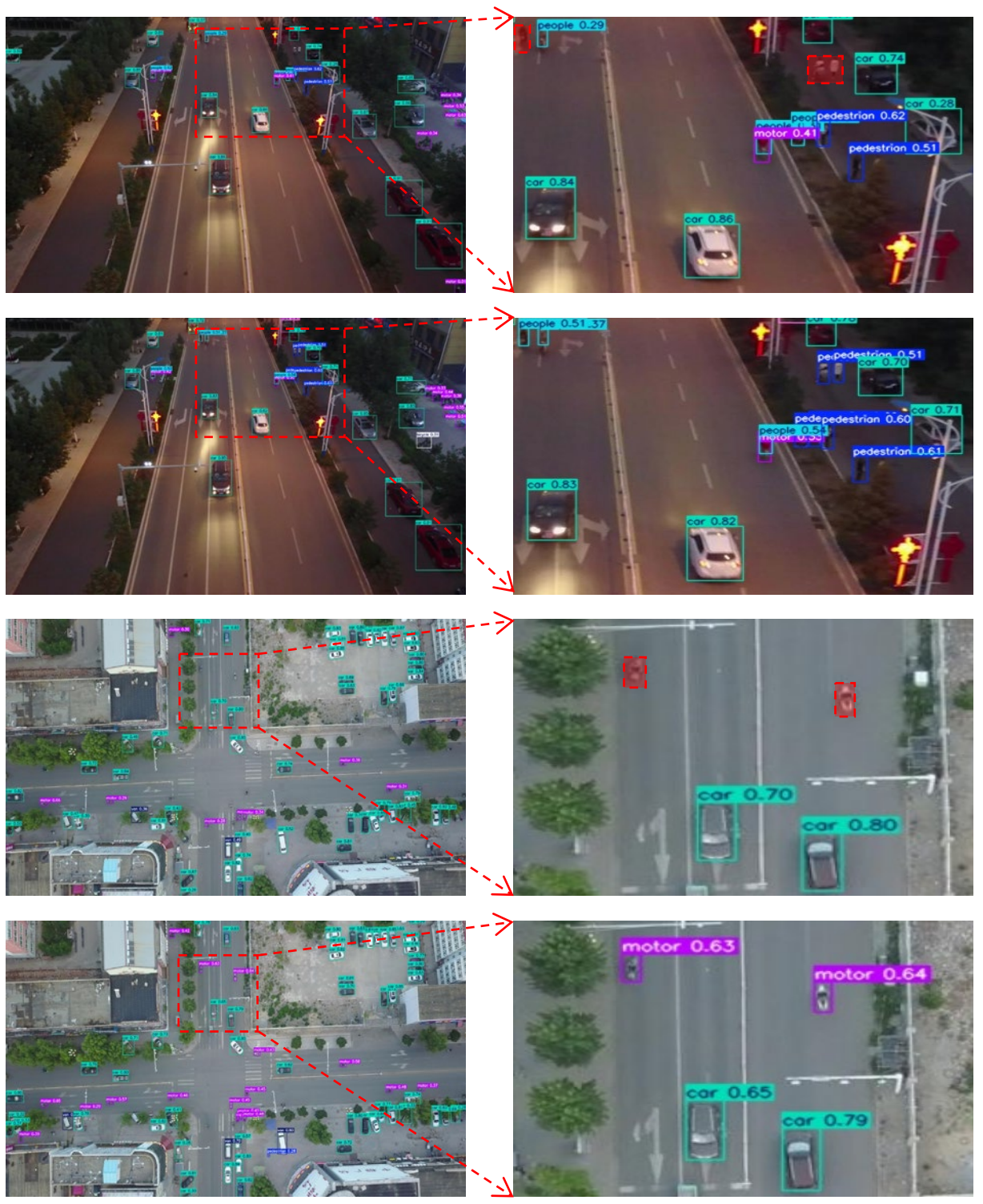

Figure 5. Comparison of YOLOv11-s (odd rows) and MASF-YOLO-s (even rows) on VisDrone2019.

## V. CONCLUSION

In this work, we improve the accuracy of small object detection in UAV remote sensing through multiple enhancements. Firstly, we introduce a small object detection layer, significantly increasing the network's ability to detect small objects. Secondly, we embed the MFAM module into the backbone network to extract rich contextual information from targets. Additionally, skip connections are incorporated into the neck network to preserve shallow semantic and reduce deep network information loss. Furthermore, the IEMA module is employed to enhance feature representation while reducing background noise interference. Finally, the DASI module is adopted to adaptively fuse low-level and high-level features, improving the feature fusion capability of the neck network. Experimental results validate the effectiveness and potential of this improved strategy, providing valuable insights for further research on small object detection.